# Addressing AI Bias in Retinal Disease Diagnostics

Philippe Burlina, PhD[1,2,3] Neil Joshi, BS[1], William Paul, BS[1],

Katia D. Pacheco, MD[4], Neil M. Bressler, MD[2]

[1]Applied Physics Laboratory, The Johns Hopkins University.

[2]Retina Division, Wilmer Eye Institute, The Johns Hopkins University School of Medicine.

[3]Malone Center for Engineering in Healthcare and Dept. of Computer Science, The Johns Hopkins University.

[4]Retina Division, Department of Ophthalmology, Brazilian Center of Vision (CBV) Eye Hospital, Brasilia, DF, Brazil.

## ABSTRACT

**Purpose:** This study evaluated generative methods to potentially mitigate AI bias when diagnosing diabetic retinopathy (DR) resulting from training data imbalance, or domain generalization which occurs when deep learning systems (DLS) face concepts at test/inference time they were not initially trained on.

**Methods**: The public domain Kaggle-EyePACS dataset (88,692 fundi and 44,346 individuals, originally diverse for ethnicity) was modified by adding clinician-annotated labels and constructing an artificial scenario of data imbalance and domain generalization by disallowing training (but not testing) exemplars for images of retinas with DR warranting referral (DR-referable) from darker-skin individuals, who presumably have greater concentration of melanin within uveal melanocytes, on average, contributing to retinal image pigmentation. A traditional/baseline diagnostic DLS was compared against new DLSs that would use training data augmented via generative models for debiasing.

 **Results:** Accuracy (95% confidence intervals [CI]) of the baseline diagnostics DLS for fundus images of lighter-skin individuals was 73.0% (66.9%, 79.2%) vs. darker-skin of 60.5% (53.5%, 67.3%), demonstrating bias/disparity (delta=12.5%) (Welch t-test t=2.670, *P*=.008) in AI performance across protected subpopulations. Using novel generative methods for addressing missing subpopulation training data (DR-referable darker-skin) achieved instead accuracy, for lighter-skin, of 72.0% (65.8%, 78.2%), and for darker-skin, of 71.5% (65.2%, 77.8%), demonstrating closer parity (delta=0.5%) in accuracy across subpopulations (Welch t-test t=0.111, *P*=.912).

**Conclusions:** Findings illustrate how data imbalance and domain generalization can lead to disparity of accuracy across subpopulations, and show that novel generative methods of synthetic fundus images may play a role for debiasing AI.

## INTRODUCTION

**Motivation:** Current deep learning systems (DLS) applied to retinal diagnostics, for diseases such as diabetic retinopathy (DR) or age-related macular degeneration (AMD), have performance approaching that of clinicians [Gulshan2016, Ting2018, Burlina2017]. This success is motivating the deployment of AI-based pre-screeners in clinical environments to address ophthalmologists' workload, and is also being used for tele-ophthalmology. However, to promote further AI insertion in retinal clinical workflows, the important problem of AI bias needs to be considered. DLSs may be affected by possible bias with regard to patients' protected attributes (e.g. race, sex, or age). AI bias may occur when training datasets are unbalanced (e.g., insufficient or no data for certain subpopulations), [Parikh2019] such as, for example, having a paucity of training images of DR for individuals of a given ethnicity, such as self-reporting as of African descent, wherein such individuals may have, on average, darker fundus pigmentation due to increased concentration of melanin within uveal melanocytes [Pakamatsu2008,Yiu2016]. Criteria for fairness can vary, and bias in AI can also have different causes, manifestations, and definitions.

**Defining AI bias:** Two definitions have received increased attention in recent work [Mehrabi2019]:

*1) (In)equality of opportunity:* consider groups of individuals with different protected attributes (e.g. ethnicity/race/origin), or simply consider subpopulations of lighter-skin vs darker-skin individuals wherein differences in fundus pigmentation, on average, might affect deep learning algorithms. An AI system designed to help plan management for AMD should not discriminate based on ethnicity/race/origin associated with darker or lighter fundus pigmentation for management recommendations if identical risk factors otherwise exist. For example, a biased healthcare recommendation system, already deployed clinically, and used for predicting individuals' need for medical services has been reported. [Obermeyer2019]

*2) (In)equality of odds:* consider groups of individuals that have different protected attributes (race, sex, age), but identical actual stages of DR (say non-proliferative DR, or NPDR), then a fair

AI diagnostic system should ascribe identical probabilities that these groups have NPDR. Unequal odds may result in unequal performance that might be affected by fundus pigmentation which influences the AI diagnostic system. Recent work in AI bias has used accuracy as a performance metric to measure bias. The criterion of unequal accuracy is adopted here to evaluate AI bias. The appendix provides rigorous mathematical expressions for the above definitions of bias and introduces a new criterion called 'delta parity'.

**Causes of bias:** Common sources of AI bias include data imbalance, distributional shift and domain generalization [Lemberger2020] whereby the DLS is faced, at inference time, with out-of-distribution samples or novel concepts when compared to training samples. This study considered a theoretic use case of data imbalance and domain generalization in a DR diagnostics DLS setting, and studied AI fairness when utilizing datasets of retinas with different appearance and markers (coloration being the principal marker, but also including other markers, as will be explained later) associated with subpopulations of individuals of lighter-skin vs. darker-skin who presumably were associated with lighter vs. darker fundus pigmentation. Such a scenario has implications for bias in ethnic/racial/origin protected factors. The next section describes prior work for addressing AI bias.

**Prior work:**  For AI bias caused by unbalanced data, a solution suggested in [Prakish2019] was to sample additional representative data for this group. When this is not practical, as is the case when AI is trained on retrospective data, and constraints exist (e.g. time, cost, regulatory limitations, logistics, or other resources) preventing such additional data collection, then other remedial approaches to AI bias are needed. This study addresses this situation and considers algorithmic de-biasing approaches. Algorithmic solutions to bias due to imbalance may include simple approaches such as using reweighting of the training loss function or more complex ones such as: using DLSs that requires less training data (so called 'low-shot method' [Wang2019, Burlina2020]); using anomaly detectors [Chalapathy2019, Burlina2019a] that work by detecting distributional inliers vs outliers; or using adversarial approaches that try to mask information about protected attributes of the individual [Mehrabi2019, Zhang2018]. Some of the aforementioned

strategies have inherent limitations. They can be impractical (e.g., adversarial training can be

unstable), unsuitable in some situations (e.g., low shot learning requires balanced samples), or

have negative effects (e.g., anomaly detectors or adversarial de-biasing solutions may

significantly decrease overall performance). Therefore, the approach used here was to use

augmentation of the training data, via generating more synthetic samples, in a controlled way, to

address the need for more data, from populations that may be under-represented, or

corresponding to missing factors. To our knowledge, this debiasing approach is novel, as is the study of AI bias in the context of AI retinal diagnostics. This approach is applied here to the use case of automated DR diagnostics but is generic and potentially could be considered for other ophthalmic, medical, or image classification uses.

**METHODS**

**Problem studied:** This study considered the use case of AI-automated binary diagnostics for

DR, i.e., non-referable (including levels 0,1) vs. referable (levels 2, 3, and 4) DR to a health care

provider -- for further follow-up or treatment (see [Cuadros2009] for definitions of these various

DR levels). It further considered the case of potential ethnic/racial/origin bias that could affect AI

due to unbalanced datasets and domain generalization. To this end, a new problem dataset was

created by altering the original Kaggle-EyePACS dataset. In this altered dataset, training data

were limited to only three subgroups of individuals consisting of specific combinations of two

factors of variations. One factor adopted here to indicate race/ethnicity/origin was taken to be the

binary factor for lighter-skin vs. darker-skin individuals (the protected attribute A) under the

assumption that this skin coloration would be reflected in the melanin concentration within uveal

melanocytes affecting fundus pigmentation. The other factor was "DR status" where we

considered the binary class labels either "referable for DR" vs "healthy" (the target $Y$). The

training data for the baseline DLS considered in this study were limited to three subgroups: DR-

referable lighter-skin (RL), no referable DR (healthy) lighter-skin (HL), and no referable DR

darker-skin (HD) individuals.[1] However, no training data were made available to the baseline DLS for DR-referable darker-skin (RD) individuals. In this domain generalization scenario, the resulting DLS would have seen examples -- and therefore learned to recognize markers -- for factors of variations consisting of ethnicity/race/origin as well as DR-status from those three groups; but then the DLS would be faced with a challenge at inference/test time, since it would also be tested on the novel combination of factors corresponding to DR-referable darker-skin individuals, in addition to the three other subgroups. Two questions are investigated henceforth: (QA) did the original data imbalance and domain generalization cause possible bias, when judged using the criterion of inequality of accuracy for the DR diagnostic DLS for darker-skin vs. lighter-skin individuals? And (QB): if there was such a bias, did our proposed synthetic data augmentation methods using generative models help debias the DR diagnostic AI?

**Dataset and labeling:** To probe questions (QA) and (QB), we created a new dataset based on the original Kaggle-EyePACS data that exhibited domain generalization by excluding referable darker-skin individuals from training but not from testing. Note that the original Kaggle-EyePACS dataset – which consisted of 88,692 public domain fundi of 44,346 participants and has been widely used by many AI researchers – was designed for ethnic diversity in mind, and does not have any inherent issues with diversity or bias (see the discussion section) [Cuadros2009].

To overcome the fact that labels for race/ethnicity/origin were not made available publicly, we added an additional annotation label for each image – focusing on a binary case of lighter-skin vs. darker-skin – rather than specific demographic definitions of race/ethnicity/origin. We asked a clinician (and co-author KP) to annotate images and classify them as one of three possible classes consisting of 1) lighter-skin 2) darker-skin 3) or "intermediate/indeterminate/unknown". Indeterminate images were not subsequently used, so as to decrease the uncertainty due to manual labeling, and only a binary value was employed thereafter. Criteria for manual labeling the images by the clinician took into consideration three factors including: A) darker vs. lighter fundus image pigmentation, since this is influenced by, on one hand, the melanin concentration within uveal melanocytes in the choroid, leading, on average, to darker fundus pigmentation, and on the other hand, the presence of melanin concentration in uveal melanocytes of darker/lighter-skin individuals. Increased concentration of melanin within melanocytes in the choroid correlate with increased melanin concentration within melanocytes in the skin [which ties it to our subpopulation partitioning]. Two additional criteria were used by the clinician selected to be influenced by markers in the retinal image that might correlate with race/ethnicity/origin. These included: B) retinal vessels appearance: individuals of African descent are reported to have larger retinal arteriolar calibers on average, [Wong2006] and C) optic disk size: larger cup to disc ratios reported in individuals of African descent. [Zangwill2004] We henceforth refer to darker-skin individuals as the groups of individuals with retinal images appearance (henceforth abbreviated as RA) that present with the above criteria/markers A)-C), and as lighter-skin individuals, those that do not. Of note, there are limitations for using these markers for the two above subpopulations as noted in the discussion section below.  Since annotation resources were limited, first a subset of 1,555 images were manually annotated by the clinician for the criteria above. We then trained a classification DLS, called the Retinal Image Appearance Extrapolation DLS (or E-RA-DLS) (henceforth we also abbreviate "retinal appearance" for the above criteria A)-C) as simply RA), to extrapolate this label to all other non-annotated images in the Kaggle-EyePACS dataset. The architecture of this E-RA-DLS was identical to the Baseline DLS described next. Last, regarding DR labels: all fundi in Kaggle-EyePACS already had annotations for DR (5 levels) which were then translated into binary labels (for referable – levels 2, 3, 4, or not referable – levels 0 and 1).

---

[1] Note: while determination of DR presence is made at the eye/retina level, we instead use the term referability – as a shorthand, but also to reflect reality – as applying to an individual.

**Preprocessing:** Each image used as input to all DLSs in this study, be they generative or discriminative, were preprocessed as follows: these were cropped to the square circumscribing the retina, zero-padded (because of the varying field of view), and resized to 256X256, when input to the generative DLSs, and 224x224, when input to the discriminative DLSs. To make the diagnostic task more challenging for the AI, no further processing was done on the data, and images that were bad-quality, or non-gradable with respect to DR, were not removed from the dataset.

**Baseline DR Diagnostics DLS (B-DR-DLS):** For this problem, as baseline, a traditional DLS for performing DR-referable classification (henceforth called B-DR-DLS) was built. Its workings mimicked approaches commonly employed in most retinal AI diagnostic studies in that it used fine-tuning of an existing network, here the popular ResNet50 [Kaiming2016, Burlina2017]. The B-DR-DLS diagnostic system was trained with the three subgroups of individuals' fundi consisting of (RL), (HL) and (HD) but no (RD), to classify referable vs. not referable DR, as explained next.

**Data Partitioning for Baseline DLS Evaluation:** The Kaggle-EyePACS fundi were partitioned into training (17,056 images), validation (4,264), and testing (400) datasets to evaluate the Baseline B-DR-DLS. The characteristic Table 1 details the partitions' sizes. Because this study aimed to keep class balance across diseased and healthy retinas, so as not to impart additional artificial bias, the combined number of training and validation fundi consisted of a total of 10,660 images for (RL), 5,330 for (HL) and 5,330 for (HD).

Testing for baseline B-DR-DLS used class balancing and included equal numbers (100 real Kaggle-EyePACS images each) for each subpopulations: (RL), (HL), (RD) and (HD), for a total of testing 400 images. Note that, to avoid introducing an additional source of uncertainty, only real images that were directly annotated by the physician for race were included in this test set, with the net effect of limiting the size of the test data. The identical test set was also used for performance evaluation of competing DLSs that attempted debiasing (detailed later), for fair comparisons. In addition, the distribution of DR levels 0 through 4 is shown, for each partition, within characteristic Table 2. The training dataset of the diagnostic DLSs included some images directly annotated for ethnicity/race/origin by our clinician, and included also images for which that label was extrapolated via E-RA-DLS (as was explained earlier).

**Metrics:** The various diagnostics DLSs were evaluated using accuracy as a main performance metric, since equal accuracy was the fairness criteria used, but also by computing sensitivity and specificity, as well as receiver operating characteristics (ROC) curves, and the area under the ROC (AUC).

**De-biasing DLSs by using Synthetic Data Augmentation via Generative Model and Latent Space Manipulation**: The goal of this study was to compare the baseline DLS, trained on a dataset affected by unbalanced data and domain generalization, to new DLSs trained on the same dataset, that was then rebalanced via the inclusion of synthetic images for the missing subpopulation (RD). Henceforth, these new diagnostic DLSs, which were then tested for their debiased characteristics, are referred as D-DR-DLSs.

Our high-level method for de-biasing used controlled synthetic data generation and augmentation and was as follows: a generative model was used to generate new synthetic images of (RD) starting with a generative system as in [Burlina2019b]. But our new generative model was also redesigned to allow fine manipulation – in latent space -- of specific factors of variations of the generated images. Specifically, this system used a gradient descent approach that started with synthetic (but realistic) images of retinas that had image markers for 'referable-DR' and transformed these – in latent space -- into synthetic images that accentuated image markers more likely associated with darker-skin individuals' (RD) retinas. This was done without changing image markers for other factors of variations, in order to preserve realism, and most importantly, preserve the referable-DR image markers. This approach is henceforth referred to as "de-biasing via altering retinal appearance" for characteristics more likely associated with darker-skin vs lighter-skin individuals. See examples of such manipulations in Fig. 1. A second method was considered, employing a latent space manipulation (gradient descent) which instead started with synthetic images of retinas with markers of characteristics more likely associated with darker-skin individuals, and transformed these images to accentuate the 'referable-DR' image markers, without altering other factors of variation, to preserve the markers for characteristics more likely associated with darker-skin individuals and maintain realism. This approach is henceforth referred to as "de-biasing via altering DR-status". The training dataset of the baseline DLS B-DR-DLS was then augmented with these new synthetic training images of DR-referable darker-skin individuals/retinas (RD) to obtain an improved training set, which contained balanced amounts of data for retinal images corresponding to the four sub-groups of individuals (HL), (HD), (RL), and importantly, (RD). More details of the processing pipeline are described next and the reader is also referred to the flow chart describing all DLSs and datasets generation and curation steps in Fig. 2.

**Additional Details for Debiasing Algorithmic Pipeline:** The approach used for de-biasing via altering retinal appearance for darker-skin vs lighter-skin individuals in synthetic images is explained next. The second de-biasing approach is implemented in a similar fashion, but instead works by altering of the image markers linked to DR. To generate more synthetic retina for DR-referable darker-skin individuals (RD), a generative model leveraging StyleGAN [Karas2018] as in [Burlina2019b] was used as a foundation. While that approach was able to generate realistic images most of the time, it was, however, not sufficient for the goals of this study. Indeed, general-purpose generative methods are sometimes prone to creating images with artifacts, such as unrealistic vascular structures, when latent space traversal is performed arbitrarily (e.g. along a rectilinear trajectory) or another process that does not precisely control the change in image markers corresponding to other factors of variation. Also, using StyleGAN for style transfer to generate more images of the kind that is missing (RD) is not possible since style transfer does not allow for precise control of image markers and instead does uncontrolled mixing. Deep learning methods that discover, control, and disentangle factors of variations in images via generative models, are desirable for this and other applications, and are currently an active area of investigation, but no method to date has definitely "solved" this challenge [Paul2020, Locatello2020].

To address this problem in DR diagnostics, the approach here works by altering retinal appearance in synthetic images, by accentuating markers tied to darker/lighter-skin individuals, while keeping markers for other factors of variation unchanged (vasculature as well as the retinal disease lesions) (See Fig. 1). The original method in [Burlina2019a] is used to start with synthetic images with markers of DR, and perform latent space manipulation on those images, via gradient descent, to generate new images that included the desired markers for subpopulations of darker-skin individuals.

To accomplish this, first a StyleGAN model was trained as in [Burlina2019a] using the same training dataset used by the baseline DLS. Pairs of (latent space vector $w$, image $I$), were then generated by using the trained StyleGAN model, in inference mode (about 120,000 $[w, I]$ tuples). Thereafter, a new retinal appearance DLS (RA-DLS), working in image space, was trained to classify between retinal images with markers of darker-skin vs. lighter-skin individuals, using the extrapolated RA labels described earlier. This RA-DLS differed from the E-RA-DLS, in that the RA-DLS is trained on more images, made up of equal numbers of fundi from darker-skin and lighter-skin individuals. Note that StyleGAN includes two latent spaces, one with an input vector $Z$ of size 512, which is mapped via a fully connected network to a new latent style tensor $w$ of size 512 and replicated 16 times for each scale for a final size of 16x512. The later latent space representation $w$ was used for manipulation of factors of variations. Subsequently, a DLS for classification of retinal appearance, and operating in latent space $w$, called the L-RA-DLS, was created, as follows: a subset of the original 120,000 synthetic images, that were classified as healthy using the Baseline B-DR-DLS, was sub-selected. We then inferred RA labels for this subset using RA-DLS, and those labels were used to train the L-RA-DLS. Finally, from the 120,000 synthetic images, a subset of 10,660 images that were classified as DR-referable by B-DR-DLS were sub-selected as starter images, and underwent the following latent space manipulation to generate new synthetic (RD) data. The corresponding latent space representations of these images*, w,* were taken, and were subject a gradient descent method to accentuate the SoftMax value of the L-RA-DLS, thereby accentuating desired image markers for individuals with darker skin. The gradient descent moved along a trajectory in latent space that was able to maximally transform images to gain the desired markers, while still keeping the vasculature, as well as the disease lesions markers, unchanged. This is in contrast with a rectilinear or arbitrary trajectory that would have produced simultaneous changes in all image markers (corresponding to ethnicity, DR status, vasculature and other factors of variations).

To formalize this gradient descent, if *(w, X)* denotes an input-output pair of latent vector w, and image X generated from the StyleGAN, where the pair was selected for *X* having a marker for DR, using the B-DR-DLS classifier, then the optimization process accentuated the retinal markers associated with darker-skin individuals, via gradient decent on *w,* as in:

$$w_{k+1} = w_k - \eta \cdot \nabla_w J(y(w), \, y=1) \qquad (4)$$

where y=1 indicates the desired probability (equal to one) for the presence of markers in retinal images X of darker-skin individuals, *y(w)* is the probability (SoftMax) output by the L-RA-DLS classifier for the input *w,* and *J(.,.)* is a loss function, taken here to be the squared difference.

**Debiased DLS creation and Evaluation**: Subsequently, a new DR Diagnostics DLS was trained, but it now also included the newly generated (RD) synthetic images in the training dataset. The number of training and validation images used now reflected parity for both race as well as DR disease status. These are also described in the characteristic Table 1. This new DLS is referred as being "optimized for retina appearance". A similar process was used to create a second debiased DLS that instead accentuated DR markers in retinal images that already possessed markers of individuals of darker-skin, to create more of the missing (RD) data, leading to a DLS deemed "optimized for DR-status".

The evaluation of these debiased DLSs was done in an identical fashion, and with the same test data, described for the baseline DLS. In addition, the improvement from baseline to debiased DLS for a specific subpopulation of (RD) was also evaluated for a set of 6,291 real Kaggle-EyePACS images with E-RA-DLS extrapolated labels using sensitivity as a metric.

**RESULTS**

Table 3 compares the accuracy, sensitivity and specificity with 95% CI for the baseline and debiased DLSs broken down by lighter-skin or darker-skin individuals.

First, the results suggests a positive answer for question (QA): As shown in Table 3, the baseline B-DR-DLS showed disparity and bias with regard to the criterion of equal accuracy across subpopulations, because it had accuracy, as defined for this study among lighter-skin individuals, of 73.0% [66.85, 79.15], vs. darker-skin individuals of 60.5% [53.52, 67.28], demonstrating disparity with delta=12.5% [3.35, 21.7] (Welch t-test t=2.670, $P$ = .008) for AI accuracy across subpopulations. For this case, we treat the accuracies as means and a debiased system as one that has equal means for the lighter-skin and darker-skin subpopulations. Thus, the Welch t-test states that, under the assumption that the system were to be unbiased, the probability of observing a statistic at least as significant as 2.670 is 0.8%,

The results also appear to suggest a positive answer with regard to question (QB): The debiased diagnostic DLSs appeared to achieve closer parity in accuracy across subpopulations. The debiased DLS operating on retinal appearance, accentuating image markers for darker-skin individuals among synthetic images of DR-referable retina (so called "retinal appearance optimized"), resulted in accuracy for lighter-skin of 78.5% [72.81, 84.19] and for darker-skin individuals of 71.0% [64.71, 77.29] with Welch t-test of t=1.729, $P$=.0847 and delta = 7.5% (95% CI: -1.0% to 16.0%). For this system, the Welch t-test states that, under the assumption that this system is debiased, the probability of observing a statistic at least as significant as 1.729 is 8.47%. Additionally, the de-biasing system relying on generating new images by accentuating image makers of DR among synthetic images with image markers of individuals of darker-skin (so called "DR optimized"), resulted in even closer parity: accuracy for lighter-skin retinal images was 72.0% [65.78%, 78.22%] and for darker-skin of 71.5% [65.24%, 77.76%] with Welch t-test t=0.111, *P*=.912 and delta = 0.5% [-8.3%, 9.3%]. Table 3 also reports the sensitivity and specificity for darker-skin and lighter-skin populations for all algorithms. In particular, the table suggests that the debiased DLS achieved improvement for darker-skin individuals, whereby the sensitivity of 35.0% [25.65, 44.35] and specificity of 86.0% [79.2, 92.8] for the baseline DLS were improved – for the "RA optimized" debiased algorithm -- to a sensitivity of 56.0% [46.27, 65.73] and specificity of 86.0% [79.2, 92.8]. The second de-biasing algorithm had similar improvements: the performance was improved to a sensitivity of 58.0% [48.33, 67.67] and specificity of 85.0 (7.0) [78.0, 92.0]. When evaluated on the leftover dataset of real data of darker-skin DR-referable individuals (RD), the sensitivity was improved from 38.48% [37.28, 39.68] to 52.63% [51.4, 53.86], for the first debiasing algorithm and of 49.75 % [48.51, 50.99] for the second, suggesting the benefit of the debiased algorithms to improve the accuracy for the protected individuals (i.e., the darker-skin individuals).

 ROC curves and area under the curves (ROC AUC) were reported for all cases in Figure 3. These also show improvement in overall parity for the debiased algorithms compared to the baseline algorithm. Finally, the DLS used to extrapolate the retinal appearance labels to the entire dataset had a validation accuracy (computed against the clinician label, and not against ground truth since that is not available) of 89.18% and 95% CI [85.17%, 93.19%].

**DISCUSSION**

**Assessing the results:** This study considered a DLS used for DR diagnostics and demonstrated that data imbalance and domain generalization for specific factors related to ethnicity/race/origin could impart unfair outcomes as measured by lack of parity in performance (which was measured via accuracy) for the baseline AI. It was applied to retinal images from presumed different subpopulations, that might affect AI interpretation of diseased retina, from lighter-skin vs. darker-skin individuals, as labeled using markers such as average fundus pigmentation (as a reflection of melanin concentration within uveal melanocytes [Wakmatsu2008,Yiu2016]), optic disc size, [Zangwill2004], and retinal arteriolar caliber. [Wong2006] Experiments suggests that the potential for bias exists in those situations, and that bias in retinal analytics needs to be carefully considered and tested in future retinal studies and DLSs that are slated for clinical deployment. This study also proposed and compared two AI de-biasing methods which may offer approaches for de-biasing retinal image diagnostic DLSs and demonstrated significant decrease in bias when measured using the accuracy performance metric.

When looking at other metrics including sensitivity and specificity: the debiased DLSs were able to achieve parity for specificity while significantly improving on sensitivity for darker-skin individuals (equal to 35.0% for the baseline DLS vs. 56.0% for the first debiased DLS and 58.0% for the second DLS). Parity was more evident when looking at the more reliable metric of ROC AUC. The ROC AUCs had differences -- between lighter-skin vs darker-skin individuals — of 0.18, for the baseline DLS, compared to much smaller differences of 0.08 for the first de-biasing algorithm, and 0.05 for the second de-biasing algorithm. Looking at ROC curves also revealed improvements: the de-bias curves largely overlap for the lighter-skin individuals, and there appears to be improvement in the ROC curves for the de-biasing algorithms for the darker-skin individuals when compared to baseline algorithm. This should be assessed, when considering, as reported by [Mehrabi2019] and others, that AI bias is usually measured vis-a-vis one specific criterion (here accuracy), and that parity often cannot be achieved with regard to several criteria simultaneously. Parity of accuracy is the most widely used metric in the bias literature, this is why that criterion was considered here. Enforcing parity for other metrics via direct incorporation into a loss function is a possible approach that is left as future work. One might also consider fairness using softer definitions of parity, as introduced in the Appendix, via the definition of delta parity. Delta was found to be 12.5% for the baseline algorithm while it was respectively 7.5% for the first, and only 0.5% for the second debiasing algorithm. The notion of delta parity opens the door to novel mathematical definitions of parity that would extend classical definitions and reflect domain-specific as well as other human-set requirements that should be investigated in the future.

**Datasets used and interpretation of the study:** It is recommended to interpret this study as being one that regards AI bias stemming from data imbalance in retinal images from groups of individuals with lighter-skin vs darker-skin – wherein the melanin concentration within melanocytes in the skin corresponds, on average, to the melanin concentration within melanocytes in the choroid that then contributes to the coloration of the retina image. However, instead of depending on just one feature of fundus pigmentation as lighter or darker pigmentation, this investigation also considered two additional features that might be associated with lighter-skin vs. darker-skin individuals, wherein groups who self-report being of African descent also, on average, have large cup to disc ratios [Zangwill2004] and larger arteriolar calibers.[Wong2006] Also, it is noted that the term race is used in this manuscript following *JAMA* guidelines (AMA Manual of Style; Section 11.12.3: Race/Ethnicity. Correct and Preferred Usage Inclusive Language). Re. terminology specifically, it is used here it in the broader context of 'race/ethnicity/origin'. This term also is used as a shorthand to indicate the attributes of a person that had the image markers and criteria A)-C) as characteristics in their retinal images. Also, regarding the data used, while this study used an experiment where a theoretical scenario of domain generalization was created by altering the original Kaggle-EyePACS dataset, it is important to note that the original Kaggle-EyePACS is itself not biased or unbalanced and that efforts were made by EyePACS to include all races, not just retinas from individuals self-identified as of African or European descent, but also included individuals self-identified as Asian, Indian subcontinent, and of American indigenous descent. Diversity of EyePACS is one of its important attributes and is the reason this dataset has been used by numerous AI and retinal disease groups for experiments.

After these experiments were completed, in personal communication with a principal that helped lead the design of the Kaggle-EyePACS dataset who evaluated this study posted on a pre-print server, our study team was made aware of statistics that clearly demonstrate that care was taken that this dataset be reflective of diversity for races and ethnicities, including self-reported African American, Asian, Whites, and other, including multiracial individuals. Based on that communication, it is noted that the dataset contains approximately 5.4% of African American individuals and about 1.3 % of referable African American individuals. Therefore, the proportion of subjects labeled as African American corresponds to a higher proportion of referable individuals compared with the population as a whole (about 24% in the dataset).

**Limitations and other remarks:** The type of distributional imbalance considered here would seem to be different from having assumed a few individuals present rather than none at all for a given subpopulation (referable-darker-skin). That scenario should be more akin to a few-shot scenario, which may be considered in the future, and related to a study such as [Burlina2020]. Also, judging from Figure 1 (b1)-(b3), the generative model, when tasked with accentuating markers for darker skin individuals, appeared to have focused mostly on accentuating pigmentation and less so on increased arterioles or venules size, or larger scleral canal opening which would result in an increased cup to disc ratio in the absence of any loss of optic nerve tissue. However, this did not seem to have prevented improvements in terms of debiasing. Furthermore, this study did not have access to the original race or ethnicity labels. The labeling was done by a single clinician (KP). It took into account several cues of A-C described earlier. Also, regarding criterion A, the retinal image pigmentation is influenced by appearance of the choroidal background influenced by the melanin concentration within the uveal melanocytes, although variations in acquisition conditions including flash, pupil size, axial length, and presence of staphylomatous abnormalities could affect this appearance as well. Labeling based on these criteria had additional limiting factors. These included: only 1,555 images were directly manually labeled, and then the labels were extrapolated (inferred, predicted in ML nomenclature) to the rest of dataset using the E-RA-DLS. This approach was motivated by constraints in annotation resources. Also, another limitation was that the E-RA-DLS was only tested against our manual annotations, and furthermore it was also used only in binary mode (excluding indeterminate labels). As such it is likely suboptimal, but still was sufficient to demonstrate debiasing power of the proposed algorithms for this proof of concept study. To avoid introducing additional uncertainty, using ground truth labels rather than extrapolated labels (which were not public at the time of this study) is another potential future endeavor, if ground truth labels for race are made available. Also, even when the ground truth for race labels become available, additional challenges still exist due to the nature of 'self-declaration' by individuals for their ethnic and race ancestry.

Note also that one could have hypothesized that bias could have originated from deeper lack of distributional shift reflected in the distribution of the granular levels of DR 0-4 across partitions. However, in examining Table 2, one finds there is similar distribution of granular DR levels for lighter-skin individuals and also with regard to darker-skin individuals for the non-referable levels (the referable levels are not represented in training by our design of experiments).

Regarding the relatively small test size: this work aimed to be diligent in the sense of only reporting on test data that were directly manually labeled by a clinician with regard to race (and not data with extrapolated retinal appearance labels). Recall that only 1,555 images were labeled manually. This fact and the constraints for balancing across race and disease labels reduced the testing dataset substantially. This study uses 400 out of 1,555 or roughly 25% of the data for testing.

The first type of debiased DLS has improved accuracy also for lighter-skin individuals (while for the second de-biasing algorithm accuracy goes down, which conforms to intuition). Regarding how this improvement could be explained for the first method:  it is possible that data augmentation played a role and that adding referable DR from individuals of darker-skin as training exemplars led to more diversity of features that also might help with performance in individuals of lighter-skin. Also, other past studies using EyePACS (e.g. [Gulshan2016]) show higher accuracy compared to our DLSs. There are various reasons for this including: differing experimental settings (such as partitioning and other factors), possible DR ground truth label noise (extra annotations by new clinicians in other studies may have been used, while here only Kaggle-EyePACS labels are directly used). Other key differences are that we report balanced accuracies across referable and non-referable classes as well as race (this inevitably leads to lower accuracy figures compared to unbalanced accuracy considering there is a strong preponderance of healthy DR level 0 retinas in EyePACS) and we also do not expunge low quality images from the dataset.

Last, while our method appears successful at de-biasing and may help in other situations of potential AI bias for retinal diagnostics, it is easy to envision cases where it may not be readily applicable. There are bias situations that may not be well addressed by this approach, and may need other solutions. This may be the case, for example, where additional real data may be needed to include the biological variability that may exist with each disease depending on a protected factor, and which may not be captured by moving on the latent space of generative models. For example, there may be different presentations of diseases in individuals of Asian vs European vs African descent, such as polypoidal choroidal vasculopathy, [Cheung2017] or image acquisition conditions may have variations depending on the protected population, different refractive errors affecting optic nerve appearances in the setting of pathologic myopia, or different media opacity across various ages that may compromise our method.

**Future work:** Future avenues for generating specific missing data and achieving better parity could use disentanglement in generative models: methods using disentangled representations [Paul2020, Locatello2019] point to the possibility of using disentanglement for debiasing. More could be done to assess this in the future. Furthermore, while this study considered lighter-skin vs darker-skin individuals as a proxy for ethnicity/race/origin and its associated potential AI bias for retinal diagnostics, other forms of bias exist including age, sex, or socio-economics that may impact AI. For example, when considering age-related macular degeneration and an epidemiologic study like AREDS, where few participants were older than 85 years, it would be of interest to investigate if AI exhibits disparity of accuracy for diagnoses in people over age 85 years, when only a few people over the age of 85 were included in the AI training set. Such investigations might be pursued in future studies.

Finally, many DLSs have shown the ability for retinal diagnostics with human level performance, in some cases, these DLSs also have used a preponderance of non-referable retina, or preponderance of individuals of a given self-reported race. These data can match the demographic makeup of a specific region or country and can be shown to work well (i.e. with regard to metrics such as accuracy or AUC) when the trained and tested on like kind data distributions. However, this success still may mask, when broken down per protected factors, such as race or age, that DLSs did not achieve equal accuracy objectives, or when faced with distributional shift, that these methods' performance degraded. In sum, and to our knowledge, the problem of bias, and more broadly of generalization and distributional shift should be given more attention in retinal AI studies. The hope is that the method presented here and other methods in [Mehrabi2019, Zhang2018] can motivate future investigations to address those issues.

**CONCLUSION**

When considering retinal diagnostics of DR, this study demonstrated that situations of data imbalance and domain generalization can affect the performance of AI diagnostics algorithms applied to a task such as DR referable versus not referable classification, and result in AI bias for individuals of presumed diverse racial/ethnicity/origins partitioned along darker-skin vs. lighter-skin populations, assuming the presumed skin pigmentation relates, on average, to the concentration of melanin within uveal melanocytes and subsequently on retinal coloration.

Our results suggest the potential benefit of certain generative methods that alter specific image markers to allow the augmentation of the diagnostic DLS and obtain parity with respect to accuracy to address this potential AI bias in retinal diagnostics.

## ARTICLE INFORMATION

### Funding

Funding: Funding was provided by The Johns Hopkins University - Applied Physics Laboratory (JHU APL) and Institute for Assured Autonomy (JHU-IAA) as well as internal grants as well as unrestricted philanthropic grants to the Johns Hopkins University School of Medicine.

### Disclaimers

The views in this paper reflect only the authors' opinion and not that of the funding entities.

### Financial Disclosures

Bressler, Burlina, and Joshi have several patents related to AI applied to retinal image analysis.

Acknowledgement: We thank Dr. Jorge Quadros (EyePACS/ Stanford University) for useful discussions about the Kaggle EyePACS dataset.

**FIGURE 1**

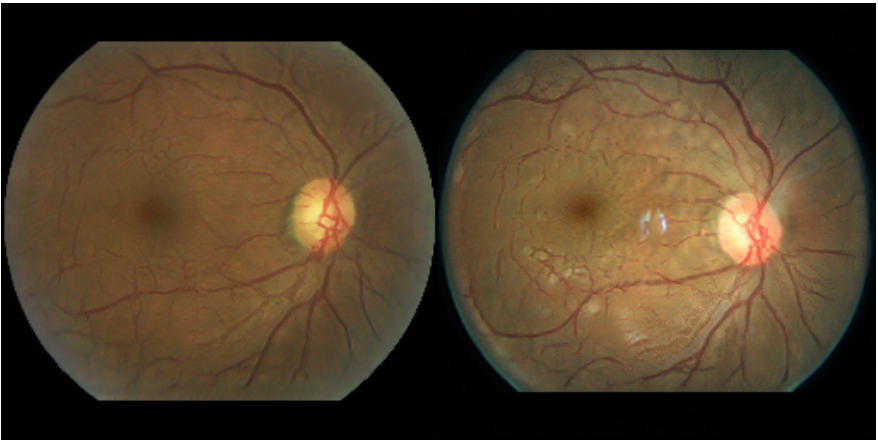

(a1)

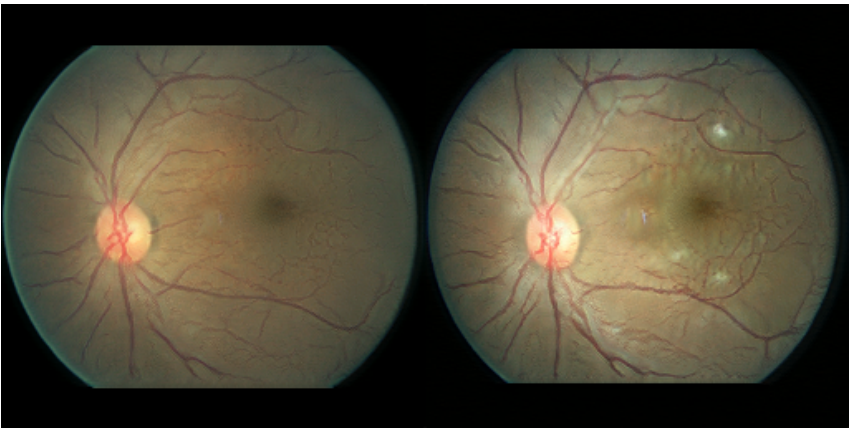

(a2)

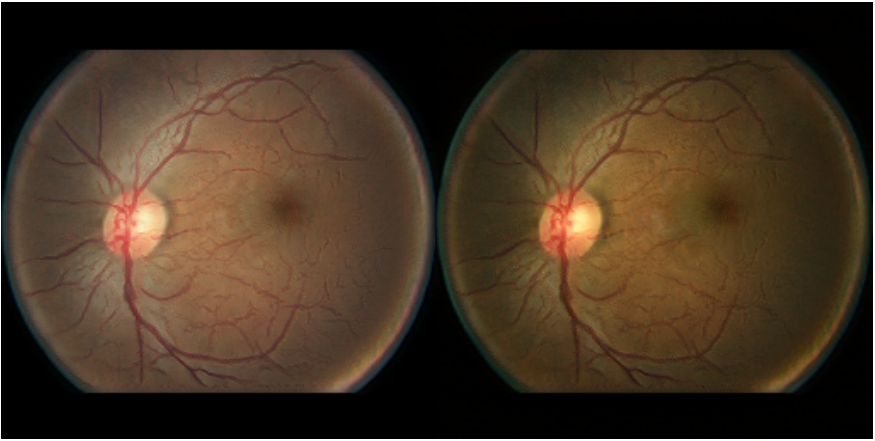

(a3)

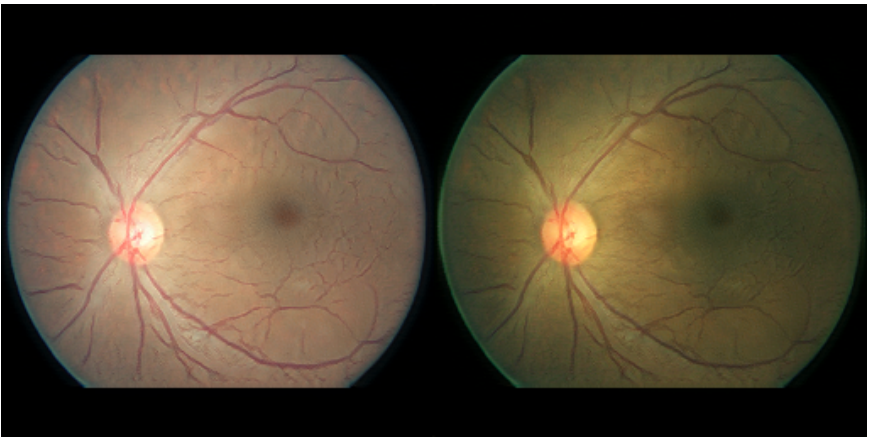

(b1)

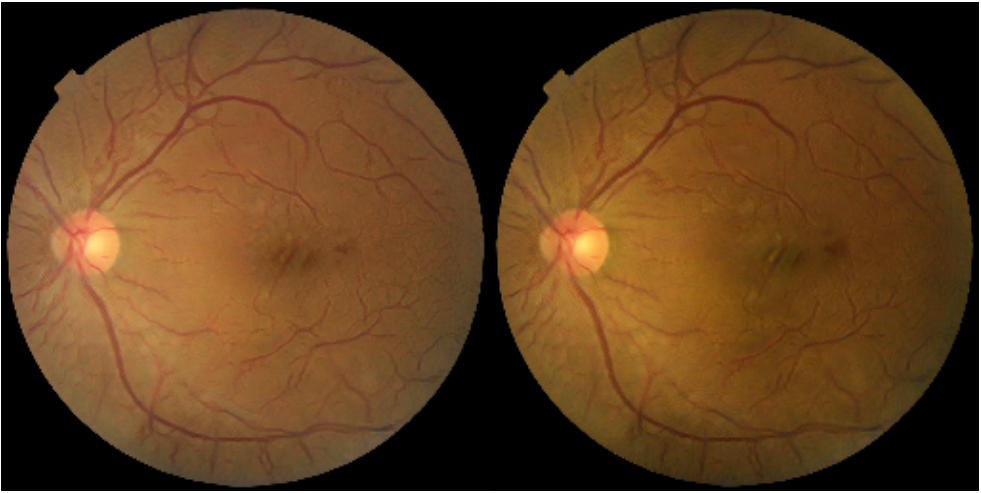

(b2)

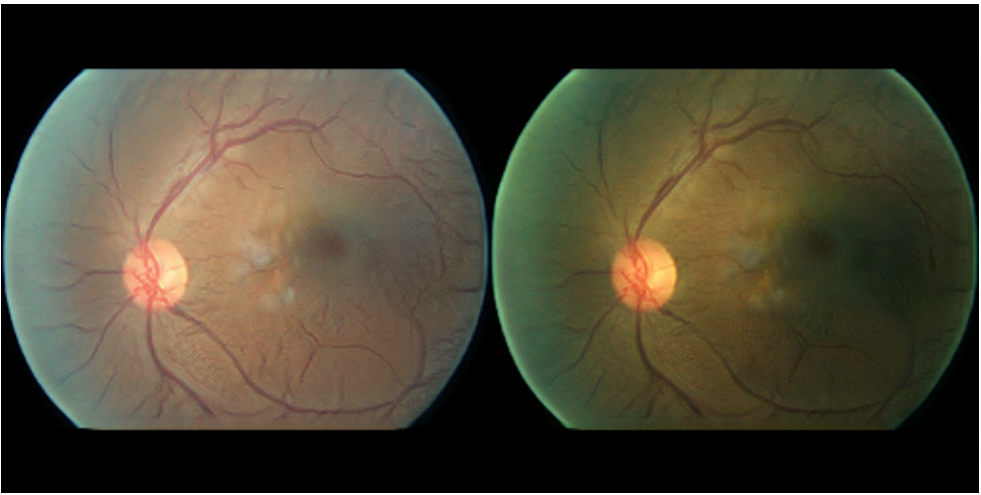

(b3)

Figure 1: The figures show right-vs-left pairs of synthetically-created images that demonstrate alterations done automatically via our generative methods, that take a synthetic retinal image as input (shown on the left), to generate a new retinal image that is of an individual with diabetic retinopathy warranting referral to health care provider and of a darker-skin individual as defined in the Methods. These generative methods are used in this study to generate images that are originally missing from the training set (i.e., of referable DR from darker-skin individuals), creating a condition of unbalance and bias.

Pairs of images in (a1)-(a3) illustrate how the proposed generative methods are used to generate new retinal images that take an input retinal image on the left, of a darker-skin individual, and accentuate the attribute 'DR-referable' in the image on the right, when compared to the left image, and leave the amount of coloration reflective of the melanin concentration within the uveal melanocytes and all other markers unchanged. The first pair (a1) starts from a retina that is not-referable but of a darker-skin individual (left image has DR level 0 or 1, i.e. no or mild DR) and converts it into one that is referable (right image is DR level 2, i.e. moderate DR) while minimally changing other attributes of the retina (the right image is also from a darker-skin individual and vasculature aspect is unchanged). Likewise, the left image pair in (a2) is of a retina from a darker-skin individual, that is not referable (left image is DR level 0 or 1) and our method then accentuates the referable attribute (right image is DR level 2) to make it referable. The same explanation applies to (a3).

Pairs of images in (b1)-(b3) instead demonstrate our complementary approach: taking as input retinal images that are already referable (left images in the pair) and altering them to accentuate the attribute 'darker-skin individuals', while preserving the DR lesions as well as the vasculature, in order to generate output images that are referable and from darker-skin individuals (right images in the pairs). (b1) in particular is already from a retina which is referable and with higher concentration of melanin within the uveal melanocytes and the method visibly accentuates melanin concentration in the right image, and both input (left image) and output (right image) have moderate DR (DR level 2). (b2) is an example where the left image is of a lighter-skin individual and already referable and our method modifies it by generating a related image of a darker-skin individual; but the method preserves the DR level, as both right and left images have visibly unchanged level 2 DR, with potential retinal hemorrhages seen. (b3) is a similar example, accentuating the left retinal image -- which is of a lighter-skin individual and referable DR, and turning it into the right retinal image of a darker-skin individual, without altering the DR level (again here both right and left images have apparently DR level 2 with retinal hemorrhages).

**TABLES**

| | Healthy*-lighter-skin (HL) | Referable-DR; lighter-skin (RL) | Healthy*-darker-skin (HD) | Referable-DR; darker-skin (RD) | Total | Healthy* (H) | Referable (R) | Lighter-skin (LS) | Darker-skin (DS) |
|---|---|---|---|---|---|---|---|---|---|
| Train (Baseline) | 5330 | 10660 | 5330 | 0 | 21320 | 10660 | 10660 | 15990 | 5330 |
| Train (Debiased) | 10660* | 10660 | 10660* | 10660** | 42640 | 21320 | 21320 | 21320 | 21320 |
| Test | 100 | 100 | 100 | 100 | 400 | 200 | 200 | 200 | 200 |

**Table 1:** Characteristic table showing the number of samples used for each population including: HL (healthy lighter-skin), RL (referable DR lighter-skin), HD (healthy darker-skin), RD (referable darker-skin), and H (total healthy), R (total referable DR), LS (total lighter-skin) and DS ( total darker-skin), and broken down by rows corresponding to the training (including training and validation) for both baseline and debiased DLS, as well as test datasets. * denotes that these are oversampled by 2 to maintain the balance of healthy and diseased factors. ** denotes that these numbers are synthetic images. Healthy* implies no diabetic retinopathy (DR) warranting referable to a health care provider.

| | Lighter-skin (LS) | | | | | Darker-skin (DS) | | | | |
|---|---|---|---|---|---|---|---|---|---|---|
| | DR0 | DR1 | DR2 | DR3 | DR4 | DR0 | DR1 | DR2 | DR3 | DR4 |
| Train (Baseline) | 4880 | 450 | 8312 | 1346 | 1002 | 4828 | 502 | 0 | 0 | 0 |
| Test | 90 | 10 | 68 | 18 | 14 | 80 | 20 | 67 | 16 | 17 |

**Table 2:** characteristic table showing the number of samples used for each population broken down by the original 5-class severity of Diabetic Retinopathy, which includes: DR0-4 for both lighter-skin and darker-skin individuals. Note that numbers for the debiased dataset are not given because there is no original 5-class grade for synthetic images.

| | Baseline DLS | Debiased DLS (retina appearance optimized) | Debiased DLS (DR-status optimized) |
|---|---|---|---|
| **Testing dataset (400 images, see Table 2):** | | | |
| Accuracy (Overall) | 66.75 (4.62) [62.13, 71.37] | 74.75 (4.26) [70.49, 79.01] | 71.75 (4.41) [67.34, 76.16] |
| Accuracy ( Lighter-skin individuals) | 73.0 (6.15) [66.85,79.15] | 78.5 (5.69) [72.81, 84.19] | 72.0 (6.22) [65.78, 78.22] |
| Accuracy ( Darker-skin individuals) | 60.5 (6.78) [53.72,67.28] | 71.0 (6.29) [64.71, 77.29] | 71.5 (6.26) [65.24, 77.76] |
| Delta-parity (signed) value | 12.5 (9.15) [3.35, 21.7]. | 7.5 (8.48) [-1.0, 16.0] | **0.5 (8.8) [-8.3, 9.3]** |
| Specificity ( Lighter-skin individuals) | 61.0 (9.56) [51.44, 70.56] | 83.0 (7.36) [75.64, 90.36] | 66.0 (9.28) [56.72, 75.28] |
| Sensitivity ( Lighter-skin individuals) | 85.0 (7.0) [78.0, 92.0] | 74.0 (8.6) [65.40, 82.6] | 78.0 (8.12) [69.88, 86.12] |
| Specificity ( Darker-skin individuals) | 86.0 (6.8) [79.2, 92.8] | 86.0 (6.8) [79.2, 92.8] | 85.0 (7.0) [78.0, 92.0] |
| Sensitivity ( Darker-skin individuals) | 35.0 (9.35) [25.65, 44.35] | 56.0 (9.73) [46.27, 65.73] | 58.0 (9.67) [48.33, 67.67] |
| **Larger Leftover Set Darker-skin individuals with DR (6291 images):** | | | |
| Sensitivity (Darker-skin individuals) (= accuracy) | 38.48 (1.2) [37.28, 39.68] | 52.63 (1.23) [51.4, 53.86] | 49.75 (1.24) [48.51, 50.99] |

**Table 3**: comparing the performance of the baseline DLS (left column) and debiased DLSs (middle and right columns) for metrics including accuracy, specificity, sensitivity and for darker-skin individuals vs lighter-skin individuals, also showing 95% error margins in parenthesis and 95% confidence intervals in brackets. Values are in %.

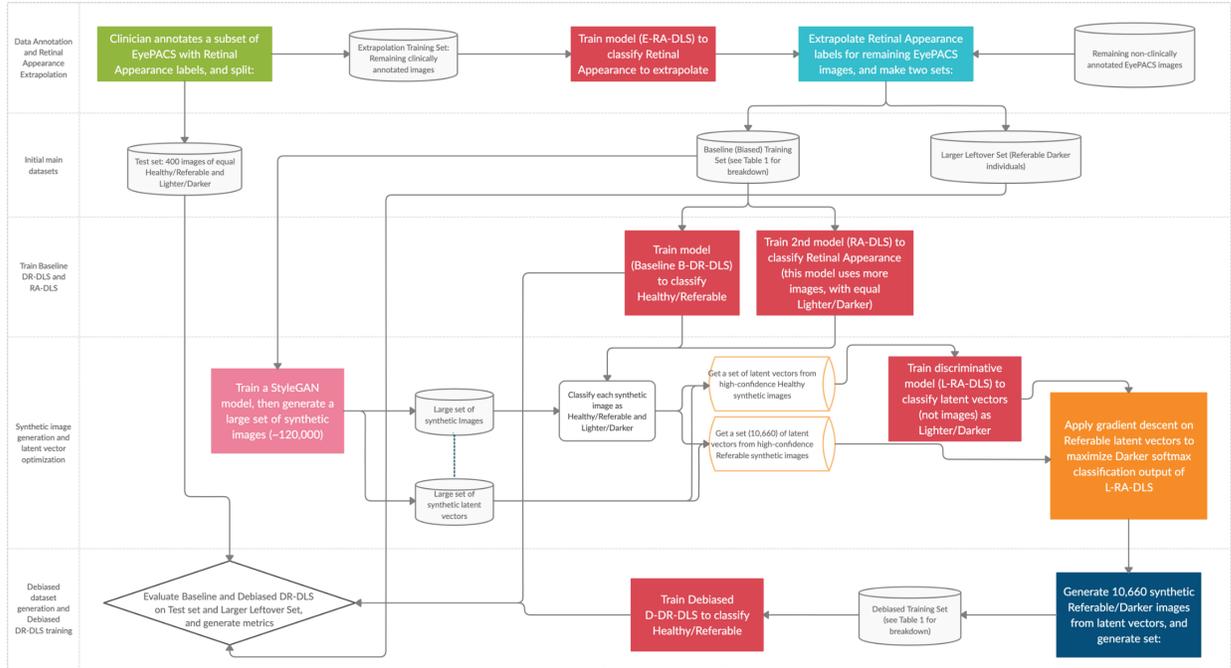

**Figure 2:** This figure details the flow chart for the debiasing algorithmic and experimental pipeline.

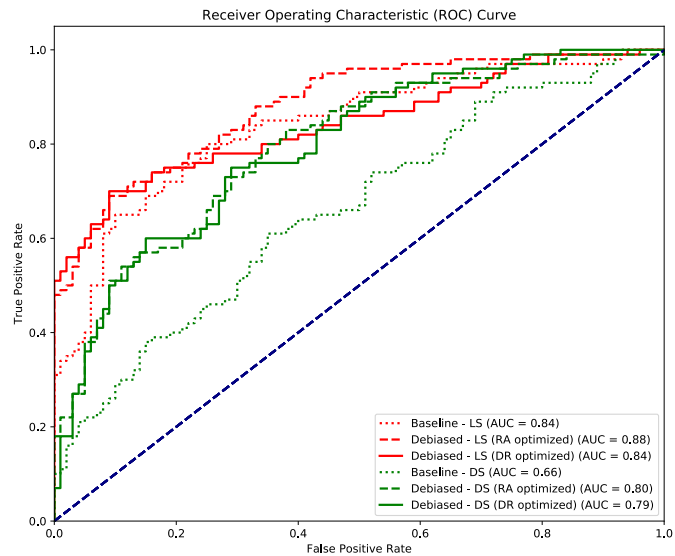

**Figure 3:** The receiver operating characteristic (ROC) curves for each population of lighter- and darker-skin individuals for both the Baseline and Debiased DLS (for both retinal appearance and DR optimized approaches). DS: dark skin, LS: light skin.

**Appendix:** Mathematical formulations of bias

Formally, the equal odds criterion in the Introduction section can be restated as follows:

For a binary protected attribute $A$ (e.g. $A$ could be such that for lighter-skin individuals $A=0$, and darker-skin, $A=1$), and a disease status $Y$ ($Y$ could represent a binary 0/1 value for the actual presence or absence of referable DR), then the AI-estimated status for the DR-referable status, denoted by $\hat{Y}$, must be independent of the attribute $A$, and must only depend on $Y$, i.e.

$$P(\hat{Y}= \hat{y} \mid A=0,\ Y=y) = P(\hat{Y}= \hat{y} \mid A=1, Y=y) = P(\hat{Y}= \hat{y} \mid Y=y) \qquad (1)$$

*for all y* [Zhang2018].

The restated formal definition for the criterion of equal opportunity makes a similar statement, i.e. that:

$$P(\hat{Y}= \hat{y} \mid A=0,\ Y=y) = P(\hat{Y}= \hat{y} \mid A=1,\ Y=y) = P(\hat{Y}= \hat{y} \mid Y=y) \qquad (2)$$

but for a given value of $y$ [Zhang2018]. Therefore, the equal odds criterion subsumes equal opportunity.

Since strict independence and equality is not practical, we also define the $\delta$ -parity criterion via the following:

$$Min \ |\delta| \ where \ \delta = P(\hat{Y}=\hat{y} | \ A=0, \ Y=y) - P(\hat{Y}=\hat{y} | \ A=1, \ Y=y)) \qquad (3)$$

*for all y* [Zhang2018]. In that case the debiasing process could be cast as one of minimizing the above objective. The problem could be to alternatively compare algorithms for debiasing based on their resulting signed $\delta$ value, a metric used herein. In this study we instead use delta as a metric for bias.